\documentclass{article}

\usepackage{times}
\usepackage{soul}
\usepackage{url}
\usepackage{placeins}
\usepackage[hidelinks]{hyperref}
\usepackage[utf8]{inputenc}
\usepackage[small]{caption}
\usepackage{graphicx}
\usepackage{amsmath}
\usepackage{amsthm}
\usepackage{booktabs}
\usepackage{algorithm}
\usepackage{algorithmic}
\usepackage{multirow}
\usepackage{todonotes}
\title{Improving Deep Learning Models via Constraint-Based Domain Knowledge: a
Brief Survey}

\date{}
\author{
    Andrea Borghesi, Federico Baldo, Michela Milano
    \\ DISI, University of Bologna
}

\begin{document}

\maketitle

\begin{abstract}
    Deep Learning (DL) models proved themselves to perform extremely well on a wide
    variety of learning tasks, as they can learn useful patterns from large data sets. 
    However, purely data-driven models might struggle when very difficult functions 
    need to be learned or when there is not enough available training data. 
    Fortunately, in many domains prior information can be retrieved and used to 
    boost the performance of DL models. 
    
    This paper presents a first survey of the approaches devised to integrate domain knowledge 
    , expressed in the form of constraints,
    in DL learning models to improve their performance, in particular targeting 
    deep neural networks. We identify five (non-mutually exclusive) categories
    that encompass the main approaches to inject domain knowledge: 1) acting on the features space,
     2) modifications to the hypothesis space, 3) data augmentation, 4)
    regularization schemes, 5) constrained learning.
\end{abstract}

\section{Introduction}

A vast array of Deep Learning (DL) approaches have been proven successful in many different learning 
tasks in recent years. One of the key strength of DL models is their ability to automatically
learn a \emph{representation} of the features composing a data set.
Deep Neural Networks (DNNs) represent the foremost and widely spread class of DL models.
Broadly speaking, DNNs are \emph{sub-symbolic} ML approaches that are very good at
extracting the useful information contained in large data sets.
One of the advantages of DL techniques is that, in general, they do not rely on 
stringent assumptions on the distribution of the underlying data and on the function to be 
learned or approximated. This allows them to be applied in many different areas with very good results,
without significant changes to the DNNs' structure and training algorithm.

However, there are contexts where purely data-driven models are not an ideal fit, for 
example when scarce data of good quality and very difficult learning tasks. 
In such situations, a great boost in the performance of neural networks (and ML models in 
general) can be obtained through the exploitation of domain knowledge, e.g. problem-specific 
information that can be used to improve the DL model and/or simplify the training process
(for instance, structured data, knowledge about the data generation process, domain experts
experience, etc).
Hence, it makes sense to take advantage of domain information to improve the performance
of DL models, so that they do not have to start from scratch while dealing
with difficult learning tasks. In other words, \emph{why learn again something that 
you already know?}.

In general, the integration of domain knowledge in DL models is a multi-faceted topic 
that has been extensively explored from multiple angles, with a wide range of approaches
dealing with different types of domain knowledge and a very large number of different 
target DL models to be boosted.
In this paper we do not claim to provide a comprehensive and exhaustive overview of all the 
methodologies proposed in the literature 
(for a broad overview \cite{von2019informed} provide a great taxonomy), 
but we want to focus on a particular class of techniques 
for injecting prior information in DL models. Specifically, we consider prior knowledge that 
can be expressed in the form of constraints and as target models to be improved with this 
information we restrict our interest to DNNs.
We take into account constraints of different nature, ranging from first-order
 and propositional logic predicates to linear and non-linear equations. In the scope of this paper the constraints can
represent relationships between the input features, relationships between input and output
features, bounds on the output variables. We consider both hard constraints (which set 
conditions that must be satisfied) and soft constraints (which set conditions with an associated
penalty, in case they are not satisfied).

The paper is structured as follows. Section~\ref{sec:related} is the core of the survey and 
presents research items from the literature for the injection of domain knowledge (expressed 
as constraints) in DNNs, grouping the related works in five macro-areas. 
Section~\ref{sec:related_areas} highlights analogies between the surveyed papers and other, closely
related, fields within the DL area. 
Section~\ref{sec:discussion} discusses common trends among the injection approaches and provides 
observations and insights. Finally, Section~\ref{sec:conclusion} summarizes and concludes the paper.

\section{Domain Knowledge Injection Approaches}
\label{sec:related}

In this section we discuss recent methods for injecting prior information in DNNs.
We assume to have a learning task where the learner is trying to approximate a function
$f^*$ that maps an input $X$ to an output $y$; the training set is composed by examples
$(x_i, y_i)$. We consider domain knowledge that can be expressed as a set of constraints, or
predicates, $\pi$. Domain knowledge can be represented by algebraic equations, such as linear and 
non-linear equations, equality and inequality constraints, logic formulas.
These predicates can be applied to different groups of variables: 1) they can involve 
only the input features $X$, e.g. it is known that the input features of well-formed data instances 
share particular properties, or are linked by a set of precise relations; 2) the constraints 
can concern only the output variables $y$, e.g. the output of the network must fall within 
a determined range; 3) the condition can involve both input $x_i$ and output $y_i$, for instance, 
the real function to be approximated $f$ could be monotonic ($x_1 \leq x_2 \rightarrow y_1 \leq y_2$).
As a particular case of constraint-based information we also include domain knowledge expressed
in the form of graphs (e.g. knowledge graphs), as graphs can be decomposed as collections
of simpler constraints encapsulating the relations between nodes and edges.
With these assumptions, the constraint-expressed domain knowledge can be integrated
in the DNNs in multiple ways. 
We classify literature works in five branches, based on the 
injection mechanism: 1) feature space, 2) hypothesis space, 3) data 
augmentation, 4) regularization schemes, 5) constraints learning. Each subsection 
is devoted to a branch.

\subsection{Feature Space}
\label{sec:features_space}

DL models' performance strongly depends on the quality of the available training data, 
either labeled or not; the features in the data define the \emph{feature space} 
whose implicit information is extracted by DNNs. The shape of the feature 
space is a critical issue for the performance of DL models and the ease 
of their training.
For instance, \emph{feature 
engineering} is a common method for improving the accuracy of 
purely data-driven ML models by selecting useful features and/or transforming the original 
ones to facilitate the learner's task. In general, this is a difficult problem and 
requires much effort, both from system expert and ML practitioners~\cite{khalid2014survey}.
In recent years, several research avenues studied the possibility to automatically 
explore the feature space in order to extract only the most relevant features, with 
most of the approaches belonging to the AutoML area \cite{nargesian2017learning}
and reinforcement learning~\cite{khurana2018feature}.
The majority of current feature engineering methods aim at selecting the optimal 
features for a specific learning task and generally tend to reduce the feature space, 
by selecting only the most relevant features (feature extraction or feature compression). 
Furthermore, these methods are generally purely data-driven and require large amounts
of data; although they aim at exploring the feature space in an efficient way, when 
the number of features is large this becomes a non trivial 
problem.

A relatively unexplored direction is the use of domain knowledge 
to create novel features that render \emph{explicit} the information hidden in the 
raw data. This is a form of feature space extension with the purpose of highlighting 
the prior knowledge embedded in the original features but not easily extractable by
a neural network. 
In practice, the approaches proposed in this area work on the original input features $X$
to generate an extended feature set $X' = X \cup \{x_j\}$, where $\{x_j\}, j \in 1,..N_j$ 
is the set composed by the additional features ($N_j$ is the number of added features).
The new features $x_j$ are computed as combination (linear or non-linear equations) 
of the original ones, depending on the domain constraints.
For example, \cite{atzmueller2017mixed} enrich the feature
space using the domain-specific information encoded by knowledge graphs. The  
relationships explicitly described by the knowledge graph represent constraints (soft and
hard) among the original input features and can be used to create additional
features, which then improve the accuracy of a supervised DNN.
In a similar fashion \cite{miao2019leveraging} increase the feature space with domain-based
 features and boost the performance of a DL model. They do not use a knowledge 
graph to obtain the additional features but rather an ensemble of decision trees solving 
a classification task on a sub-set of the data devoted to the training of the DL model. Each 
tree learns domain-specific information and partakes to the final prediction 
through its own score; these scores are then added as additional features to the 
training set. 
Another method to incorporate domain knowledge by enriching the feature space is discussed by
\cite{berrar2019incorporating}, which study the improvement of a DNN for the prediction
of soccer match outcomes obtained through the addition of domain-inspired features. The training 
set is a time series composed of matches among two different teams and related outcomes; 
their approach consists in adding a set of novel features that encapsulate i) the rating of the 
teams involved in the match and ii) the results obtained in the last $k$ matches by each team.
These novel features are encoded as a set of linear and non-linear equations applied to
the original input $X$.

\subsection{Hypothesis Space}
\label{sec:adhoc_net}

A DNN can be characterized by its position in the so called \emph{hypothesis space}, 
namely the multi-dimensional space covering its structure and its hyperparameters. The architecture of a NN 
is a very relevant factor for determining its performance on different learning tasks.
The hypothesis space has been explored with implicit guidance provided by domain knowledge 
 for many years, as attested by the introduction of convolutional networks, whose structure is based 
on the locality assumption (e.g. pixels close to each other in an image are related). 
Implicit knowledge about temporal locality has also lead to a wide 
range of architectures targeted at handling time series and sequences, for instance recurrent NNs (RNNs), 
Long-Short Term Memory NNs (LSTMs), Temporal Convolutional Networks~\cite{bai2018empirical}.

In more recent years, a remarkable research effort has been devoted to exploring DNN 
architectures optimized for circumstances where the domain knowledge can be expressed 
in the form of graphs, precisely with a DL model called \emph{Graph Neural Network} (GNNs) 
\cite{scarselli2008graph}; \emph{Graph Convolutional Networks} (GCNN)~\cite{Kipf:2016tc} 
 were introduced as well to exploit the same type of graph-expressible prior information.
GNNs have
been used in several fields~\cite{zhang2019graph}, owning to their capability
to deal with data whose structure can be described via graphs, thanks to a generalization 
in the spectral domain of the convolutional layers found in many deep learning networks. 
GCNNs most common applications involve semi-supervised classification tasks, with the goal
of predicting the class of unlabeled nodes in a graph -- a case of graph learning.
%However, graph convolutions have been applied to disparate learning task from several areas,
%from image classification\cite{liang2018symbolic} to physics predictions\cite{mrowca2018flexible}.
Using prior information expressible as graph has been proposed also to devise 
other types of NNs, namely networks whose structure resembles the LSTM's one but where the 
nodes connections are determined by the prior information~\cite{marino2016more}. Similarly,
\cite{jain2016structural} combine the temporal structure of RNNs with spatial-based information,
namely the domain information is encoded as sequences of actions each one characterized by a time
and a position in space, which are then represented through an extended RNN.
As mentioned earlier, domain knowledge represented as a graph can be expressed by sets of constraints,
which encode the relationships between the nodes through the edges. 
Currently, the vast majority of information injection methods in this area 
(and all those listed in this section) operate on the input features $X$, which can 
be cast as concepts and related connections within the graph structure.

\subsection{Data Augmentation}
\label{sec:data_augmentation} 

Besides working on the feature and the hypothesis space, another mechanism to infuse domain 
knowledge in DNNs regards the training data, mainly in the forms of creating \emph{ex-novo} 
entire training sets or adding new examples to 
existing ones following criteria defined by the domain knowledge (for instances,
examples respecting certain relationships among the input features).
We refer to these methodologies with the term \emph{data augmentation}.
Data augmentation approaches based on prior information (constraints) have been
started to be explored in recent years, especially to cope with data sets 
of limited size and the related issue of poor generalization performance~\cite{hernandez2018data}.
%\cite{hernandez2018further}
Data augmentation techniques have a strong history of success in the context of image-based 
learning tasks (e.g. image classification)~\cite{wang2017effectiveness}. For image 
classification tasks, the training set can be
augmented by applying a plethora of transformations to the images in the original 
training set, thus feeding the NN to be trained
with a more varied set of examples. The selection of the best transformations to apply
to augment the available data is a process that is typically guided by information 
obtained via domain experts. In particular, constraint-based domain information can
be used to augment the available data, e.g. linear and non-linear functions such as
rotation, distortion, flipping, etc.
For instance, \cite{bjerrum2017smiles} propose a data augmentation technique to train
a LSTM model for classifying chemical molecules. Each molecule can be described as 
a combination of its composing elements, encoded as a concatenation of strings. A single 
molecule has multiple possible encoding strings; the authors propose to expand
the training set by enumerating the chemically allowed combinations for each data 
point/molecule. The enumeration takes place via a heuristic algorithm that enforces 
on the generated examples the same chemical properties of the original data points;
these properties are encoded as a collection of constraints (linear combinations representing
admissible chemical properties) among the input features (the 
concatenated strings). In this case the constraints involve both input features $x_i$ 
and the output $y_i$, as the relationships among $X$ are used to generate novel
data points with the same output value (the label).

\cite{mollaysa2019learning} propose a different methodology for data set augmentation: they
consider \emph{feature side-information}, domain knowledge describing feature properties 
and/or feature relations. The feature side-information are expressed as a matrix whose
rows represent the prior information associated to each feature; a similarity function is 
introduced to compute the pairwise similarity of different features. 
An augmented training data point is obtained from an original example by applying a 
transformation that preserves the associated label and perturbs the values of similar 
features. Again, the focus is on constraints among input features $x_i$.
Similarly, \cite{vo2017combination} devise a data augmentation method based on 
similarity score among features, for the purpose of improving the results of a CNNs used
for sentiment analysis. In this case the original data set is composed by labeled 
sentences (the training examples); each sentence can be decomposed in a set of sentiment
terms, i.e. the features. The author introduce a similarity measure for the sentiments and 
then propose an algorithm (based on quadratic programming) to generate similar sentences
from the original ones, based on the sentiment similarity score; the augmented 
similar examples are annotated with the same label as the corresponding original 
training points.

\subsection{Regularization Schemes}
\label{sec:regularization} 
Regularization is a widely known method to avoid overfitting in machine learning models, 
but it can also be exploited to inject domain knowledge. To 
do so, some form of prior insight, i.e constraints,
is translated into a regularization term able to measure the level of consistency with the domain knowledge and guide the loss optimization process. More in general, the loss function can be defined
as follows:
\begin{equation*}
    \sum_{i=1}^N \lambda_{\tau} L(f(x_i), y_i) + \lambda_{\pi} L_{\pi}(f(x_i))
\end{equation*}
where $L$ represents the true loss, e.g.~MSE, while $L_{\pi}$ represents the regularization term
given a set of constraints $\pi$; these terms are weighted, respectively, with $\lambda_{\tau}$ and 
$\lambda_{\pi}$; the choice of these weight parameters is non-trivial and might consistently affect the final outcome
(\cite{fioretto2020lagrangian} address this problem exploiting \emph{lagrangian duality}).
In the last few years many authors worked on this class of methods proposing a variety of
approaches to the problem; for instance, \cite{muralidhar2018incorporating}, developed a DNN, 
called \emph{domain adapted neural network} (DANN), which exploits a mathematical formulation of the constraints
, namely approximation and monotonicity constraints, whose degree of satisfaction
is used as regularization term.
A more general approach is introduced by \cite{diligenti2017semantic} 
with \emph{Semantic Based Regularization} (SBR), which provides a set of rules
to translate first-order logic formulas into fuzzy constraints, then used as penalty factors in the loss function. This is achieved by introducing a \emph{t-norm} function,
which can be defined in different ways, according to the desired interpretation of 
the domain constraints involved in the regularization, allowing for an adaptable tool.
\cite{marralyrics} present a strongly related method denoted \emph{LYRICS}, a 
framework that introduces a declarative language to express the domain knowledge and 
enforces it using SBR on top of DNNs,
allowing a very flexible representation of the constraints and the prior information. 
A stochastic approach with semantic loss is proposed by \cite{xu2017semantic},
which use a regularization term given by the probability of generating a state 
satisfying the domain-based constraints; during the training process, the presence of states
not satisfying the desired constraints is penalized acting on the loss function.
\cite{silvestri2020injectDomainKnow} propose a SBR-inspired technique for injecting 
domain constraints in a DNN used to extend a partial variable assignment for the Partial
Latin Square problem, specifically finding feasible solutions that respect domain constraints.

%Another relevant approach is provided by \cite{fischer2019dl2} with \emph{DL2}, a network
%architecture allowing to query the ML model to very the satisfaction of domain constraint 
%given a specific input and to train the network itself using the prior knowledge expressed 
%in a declarative way. 
%\cite{fischer2019dl2} present
%a method for translating logical constraints in loss functions that guide the training 
%towards the desired output. \cite{muralidhar2018incorporating} propose a different approach
%to incorporate domain knowledge in a NN by adopting a loss function that merges
%mean squared error and a penalty measuring whether the NN output respect a set of constraints
%derived from the domain. \cite{xu2017semantic} introduce a method to integrate semantic 
%knowledge in deep NNs, again exploiting a loss function; in this case the approach is 
%targeted at semi-supervised learning and not well suited for supervised tasks.
%Acting on the loss function with a regularization term has been proposed also by
%\cite{diligenti2017semantic}, with their work on Semantic-Based Regularization (SBR), 
%a method to merge high-level domain information expressed as first-order logic in ML models.

\subsection{Learning with Constraints}
\label{sec:constraint} 

DL models are often employed in specific tasks, e.g. to find solutions to optimization problems. 
Usually these approaches follow a two-step
process where: first the model is trained with the observed data and then used to approximate
an aspect of the optimization problem, e.g. the cost function. In these cases, the model's 
accuracy is not the only way to measure the performance of the approach, as task related
metrics might be more relevant.
Lately a new paradigm emerged, namely \emph{decision-focused learning} (or \emph{end-to-end learning}),
where DNNs are trained to directly produce good results for the end goal of the whole task. 
In this context, prior information for the specific optimization task is introduced to
guide the training of the DNNs used to approximate the needed functions.
This class of approaches shares some aspects with the regularization ones, as the penalty 
terms in the loss function can be applied also for end-to-end learning. 
However, since in this case the DNNs are used within larger optimization models, the mechanism to
influence their behaviour is different. In general, the DNNs for decision-focused learning
produce a proxy (intermediate) solution for the target task, given a fixed parametrization of the network,
then optimized by the loss function; however, as the solution space might be not continuous, differentiability
issues might arise, which are in general solved through transformations.

%depend on the results of the target problem, 
%this requires to produce a proxy solution of the end task given a fixed 
%parametrization of the neural network; this approach however introduce issues of differentiability, 
%since the solution space might be not continuous; usually some form 
%of transformation is introduced to enforce this property.

A first example of this method can be seen in \cite{wilder2019melding}, where it
is applied to combinatorial optimization with the introduction of a continuous relaxation of the 
proxy solution (which is discrete), namely a \emph{convex hull}; the loss 
is then computed by combining the gradient with respect to the decision variables and 
the gradient of the model with respect to its own parameters. 
%In \cite{wilder2019end} the focus is 
%on graph optimization, especially on a link prediction problem; in this case the differentibility is 
%achieved with a specific architecture of the network (called \emph{ClusterNet}), where through a 
%\emph{graph embedding}, performed using a GCN, the graph representation is mapped into a continuous
%space and then used to produce the proxy solution. 
%\\
A slightly different contribution is given by \cite{donti2017task}, which apply end-to-end learning 
on a stochastic optimization problem; however, in this approach the focus is on the training 
process, where the gradient descent updates the network parameters using two functions: 
one computing the number of constraint violations and one represented by a classic loss function.
If the proxy solution does not satisfy the constraints, the parameters are 
updated using the first function, otherwise using log-likelihood loss. \\
\cite{li2019augmenting} offer a totally different method to tackle problems where 
prior knowledge is expressed with first-order logic formulas; in this case the
prior information is embedded directly in the DNN, where nodes, 
called \emph{named neurons}, are labeled to mimic the logic elements in the formula; these are used
to build \emph{constrained neural layers}, which produce their output according to
the truth value of each named neuron in the layer. \\
Recently, \cite{fischer2019dl2} introduced an approach similar to the end-to-end learning paradigm, 
but more general, called \emph{DL2}.
DL2 is a framework for explicitly embedding constraints in DNNs, specifically, it allows to translate
logical formulas into loss functions, i.e. by defining recursively the corresponding mathematical
equation for each term in the formula; the training is then carried out using projected gradient descent.
The model is also provided with a SQL-like query language, 
which allows to interrogate the network on a specific input, in order to:
check if the constraints are satisfied and train the model for input outside of the set of observed data;
basically, the network can be challenged with queries that help improve its accuracy
beyond the data available for the training -- this method is called \emph{global training}.

\section{Related Areas}
\label{sec:related_areas}

Another field where prior information has been exploited to improve performance is
\emph{reinforcement learning}. In this area as well the knowledge domain information can be used 
to improve the models and/or transform very difficult problems into more tractable ones;
for example, by creating good initial conditions for the training algorithm, 
thus decreasing the number of required training examples and therefore providing a 
\emph{warm start} for the reinforcement learning process~\cite{silva2019prolonets}. 
Prior information has been successfully applied with remarkable benefits 
 in various context well suited for reinforcement learning~\cite{luketina2019survey}.
However, an important aspect has to be considered while injecting domain knowledge in 
this setting: uncertainty in the domain knowledge can greatly hinder the learning
process if wrong decisions (caused by uncertain or incomplete information on the system state)
are taken at the beginning of the learning phase~\cite{terashima2011study}.
Another area where domain knowledge has been shown to be beneficial is the initialization
of the weights of a DNN.
For instance, \cite{yuan2020novel} describe a semi-supervised pre-training strategy 
for DL models to predict the behaviour of industrial processes; the pre-training is based on domain 
information regarding the chemical properties and relations among the data set features.
With a similar strategy \cite{gulccehre2016knowledge} exploit semantic-based knowledge to address an artificial
task unsolvable by standard ML techniques; the key idea is to provide ``hints'' to the learner
about appropriate intermediate concepts.

\section{Observations \& Remarks}
\label{sec:discussion}
In this section we will discuss some common traits spanning over
all the knowledge injection methodologies described previously and present some insights.

\paragraph{Feature manipulation} 
The features composing the data are often the target of the techniques described in Sec.
~\ref{sec:related}, as they offer a practical mechanism to inject prior information in a 
variety of DL models. This method is especially effective when the knowledge can be 
expressed in the form of relationships among input features, which can then be used either to 
augment the number of training examples (the relations provide a guide to generate novel 
valid examples starting from the original ones) or to create new, more informative features 
that should highlight implicit information already available in the data but hard to extract.
An interesting thing to be noted is the fact that acting on the feature space does not 
necessarily require information about labels $y_i$, as the domain knowledge might involve
only constraints among the $x_i$ features; this suggests that this type of approach is
especially well suited for unsupervised learning tasks.
These approaches go against the most prevailing DL research direction in recent years, 
namely the development of purely data-driven models that extract all the hidden information
contained in the feature space without any help. Clearly, having powerful and general models
capable of good performance regardless of the domain information is a very important
purpose, nevertheless we reckon that there exist many contexts that could benefit from the 
exploitation of domain-derived prior information, and this research direction is worthy 
to be explored in future works.

\paragraph{Accuracy VS Optimization}
A partially unexplored area, in our opinion, is the trade-off between the accuracy of the 
DL models (e.g. high accuracy for classification tasks and low error for regression ones)
and the satisfaction of the constraints imposed to the models. For instance, we already
stated that in \emph{end-to-end} learning the goal is to optimize the neural network
given a specific task, therefore we might not observe an improvement in the accuracy of
the model itself (e.g. Mean Average Error), but rather an improvement over a specific
task. This is also the case for regularization schemes, where the resulting models have
 outputs more consistent with the prior knowledge, but with no increase in accuracy in mere 
terms of prediction. This is an aspect that is implicit in the desire to minimize a 
objective function composed by multiple terms that do not point towards the same direction.
Generally speaking, the majority of these approaches opt for ``soft'' constraints
on the output of the neural network, that is the constraints are not enforced strictly
(as in the case of hard constraints) but rather the optimization tries to balance the 
diverse terms.
This problem has been only partially studied on DNNs \cite{marquez2017imposing,detassis2020movingTargets}, 
although in the optimization area it is  well known that minimizing an objective function
with multiple terms yields poor convergence properties, mainly because the optimizer is 
likely to focus on one term of the objective function while ignoring the remaining ones;
furthermore, enforcing multiple constraints of different natures means that the terms 
risk not to be commensurate. Finally, an additional complicating factor in these methods is the 
selection of adequate weights for the various terms (model accuracy and constraints-based
terms).

\paragraph{Small data sets}
A non-negligible problem for DL approaches is the requirement of large amounts of data, 
preferably labelled. In many scenarios, this does not happen, thus the training 
of deep models is hindered. Injecting domain knowledge can boost the performance of DL 
models when training data is scarce. For this purpose, the most obvious candidates are
techniques which augment the available data, but regularization schemes and feature 
engineering approaches can provide benefits as well, by, respectively, ``guiding'' the training 
process of the DNN and simplifying the learning task thanks to the additional features
added to the raw data. In general, techniques to integrate prior information can 
be extremely useful in the context of \emph{active learning} and other settings where 
the dearth of data cannot be bypassed. Nowadays, the majority of approaches for active learning
are not guided with domain knowledge whilst there are potentially big benefits in 
exploiting such knowledge, for instance to drive the selection of new instances to be
evaluated, an exploration that is currently guided by domain-agnostic strategies based on
measures such as information gain. Some recent works did preliminary work towards this 
direction. For instance, \cite{borghesi2020smartFPtuner} propose a combination of active 
and end-to-end learning, by embedding a DNN within an optimization model for floating point variables
precision tuning. As the training set is relatively small, the DNN starts with inaccurate
predictions; the author use active learning to iteratively improve
the DNN by retraining it on new examples, namely the solutions of the optimization 
model, directly depending on the domain knowledge.

\paragraph{Evaluation metrics} 
A partially unexplored issue in the domain information injection area regards the best 
metric to evaluate the performance of the injection mechanism. Broadly speaking,  
most common DL techniques are measured on the basis of a single metric such as accuracy 
(classification tasks) or Mean Average Error or Mean Squared Error (for regression tasks).
However, these metrics are not the fairest ones when one has to judge the benefits of 
injecting domain knowledge. For instance, a common aspect of regularization schemes and methodologies
that embed constraints in the NNs is that their goal is not to simply reduce the prediction 
error or increase the model's accuracy, but rather to obtain NNs whose output respects
some desirable proprieties (e.g. monotonicity) or where certain relationships between input and
output need to hold. In practice, this means that there are no established and straightforward
methods to measure the improvements of knowledge injection methods, as the for each domain
different authors chose different evaluation metrics. In many cases the authors compare 
their injection methods to standard DL models using specific test sets, carefully crafted 
for the particular task. For example, models enforcing an output with no violations
of a particular constraint can be trained on data sets that contain instances violating 
the constraints but tested exclusively on data sets with no violations. This is acceptable
in order to perform a fair comparison but with an increased risk of creating
 artificial experimental settings.
This lack of homogeneity is not a trivial issue and it complicates
the comparison of different techniques. A set of common benchmarks and key performance indicators would greatly benefit 
this research area.

\paragraph{Towards a unified framework}
As an overall remark, it can be noted that the various approaches discussed previously 
operate in relative isolation. This is understandable and can be partially explained by the
fact that domain injection techniques are, by definition, domain specific. This issue is
exacerbated by the large number of possible injection mechanisms and targets.
This leads to 
a lack of a common perspective and makes the comparison of different injection approaches harder;
however, some recent attempts have been made, see for example \cite{borghesi2020injectIJCAI}, where
multiple knowledge injection techniques are employed to boost a DNN dealing with a complex 
learning task.
We believe that a unified methodology, or framework, for injecting domain-derived constraints
would be a great step for advancing the research progress in this area. 
Such unified framework would require a common language to express the domain knowledge; it should
be a language with expressive power (many different types of relationships and concepts should be
allowed to be represented) and flexible, that is capable to describe information stemming from very diverse domains.
We reckon that a language based on constraints and logic predicates could be a very apt choice for this task, especially
thanks to paradigms such as constraint programming and mixed integer and linear programming, which
have been proven to be capable of handling a wide range of domain-specific challenges. 
The next step would be deciding the best information injection mechanism for the desired task; for
a detailed answer follow-up studies need to be conducted, but some guidelines can be already provided
at this stage.
First, when the available data is scarce, data augmentation and feature space manipulation 
techniques are to be preferred, as they allow to increase the training set size by
exploiting known relationships between input set features $x_i$, as well as relations between input
features and output features $y_i$. 
Secondly, if the information contained in the input data is 
hidden and not easy to be extracted, ad-hoc DNNs architectures can be extremely helpful -- e.g. acting
on the hypothesis space -- as the domain expert knowledge can be directly injected in the 
neural network structure from its design to the training algorithm. 
Finally, regularization methods and explicit constraints learning in the neural network are 
very effective strategies for end-to-end learning and, in general, for more complex learning 
tasks where the model accuracy is not the exclusive performance metrics. In this case, careful 
attention should be employed to choose the right trade-off between the optimization problem 
and the pure minimization of the DNN's loss function. This is a non trivial issue, as it complicates
the actual development of regularization and constraints learning techniques, together with 
the fact that they are extremely task-specific (e.g. a set of weights for one context would be
not well suited to other ones); the 
practitioner implementing these approaches will be faced with steeper challenges compared, for
example, to data augmentation approaches, which also tend to be more transferable.

\section{Conclusion}
\label{sec:conclusion}

The integration of domain knowledge expressible in the form of constraints into deep neural networks
is a wide research area that have seen increasing research interest in recent years. 
This topic has been tackled from different angles by a variety of approaches, typically
in relative isolation, a fact that probably hindered research breakthrough.
In this paper, we have provided a first cross-disciplinary attempt at classifying existing approaches,
identifying the main classes of techniques for domain knowledge injection, and highlighting
connections with related fields from the DL area. Moreover, we identified a series of common
trends and issues that have been addressed and open challenges that still need to be tackled, 
with the hope of providing useful insights and a guidance for future research efforts.

\section*{Acknowledgments}
This work has been partially
supported by European H2020 FET project OPRECOMP (g.a. 732631) and ICT project AI4EU (g.a. 825619).
\FloatBarrier

\bibliographystyle{abbrv}
\bibliography{bib}

\end{document}